\documentclass[letterpaper]{article} % DO NOT CHANGE THIS
\usepackage{aaai2026}  % DO NOT CHANGE THIS
\usepackage{times}  % DO NOT CHANGE THIS
\usepackage{helvet}  % DO NOT CHANGE THIS
\usepackage{courier}  % DO NOT CHANGE THIS
\usepackage[hyphens]{url}  % DO NOT CHANGE THIS
\usepackage{graphicx} % DO NOT CHANGE THIS
\usepackage{natbib}  % DO NOT CHANGE THIS AND DO NOT ADD ANY OPTIONS TO IT
\usepackage{caption} % DO NOT CHANGE THIS AND DO NOT ADD ANY OPTIONS TO IT
\usepackage{algorithm}
\usepackage{algorithmic}
\usepackage{booktabs}
\usepackage{multirow}
\usepackage{xcolor}
\usepackage{pifont}
\usepackage{amssymb}
\usepackage{amsmath}
\usepackage{makecell}
\usepackage{xcolor}
\usepackage{subfig}
\usepackage{makecell}
\usepackage{amsthm}
\newtheorem{theorem}{Theorem}
\usepackage{newfloat}
\usepackage{listings}
\DeclareCaptionStyle{ruled}{labelfont=normalfont,labelsep=colon,strut=off} % DO NOT CHANGE THIS
\floatstyle{ruled}
\newfloat{listing}{tb}{lst}{}
\floatname{listing}{Listing}
\title{Transferable Graph Condensation from the Causal Perspective}
\author{
    Huaming Du\textsuperscript{\rm 1}, Yijie Huang\textsuperscript{\rm 1}, Su Yao\textsuperscript{\rm 2}, Yiying Wang\textsuperscript{\rm 3}, Yueyang Zhou\textsuperscript{\rm 3}, Jingwen Yang\textsuperscript{\rm 3}, Jinshi Zhang\textsuperscript{\rm 3}, Han Ji\textsuperscript{\rm 3}, Yu Zhao\textsuperscript{\rm 1}, Guisong Liu\textsuperscript{\rm 1}, Hegui Zhang\textsuperscript{\rm 4}$^*$, Carl Yang\textsuperscript{\rm 5}, Gang Kou\textsuperscript{\rm 6}\thanks{Corresponding Author.}
}
\affiliations{
    \textsuperscript{\rm 1}Southwestern University of Finance and Economics\\
    \textsuperscript{\rm 2}Tsinghua University\\
\textsuperscript{\rm 3}Ant Group\\
\textsuperscript{\rm 4}Dongbei University of Finance and Economics\\
\textsuperscript{\rm 5}Emory University\\
\textsuperscript{\rm 6}Xiangjiang Laboratory\\
dhmfcc@swufe.edu.cn, yaosu@tsinghua.edu.cn, hegui.zhang@qq.com, kou.gang@qq.com
}

\usepackage{bibentry}
\begin{document}

\maketitle

\begin{abstract}
The increasing scale of graph datasets has significantly improved the performance of graph representation learning methods, but it has also introduced substantial training challenges. Graph dataset condensation techniques have emerged to compress large datasets into smaller yet information-rich datasets, while maintaining similar test performance. However, these methods strictly require downstream applications to match the original dataset and task, which often fails in cross-task and cross-domain scenarios. To address these challenges, we propose a novel causal-invariance-based and transferable graph dataset condensation method, named \textbf{TGCC}, providing effective and transferable condensed datasets. Specifically, to preserve domain-invariant knowledge, we first extract domain causal-invariant features from the spatial domain of the graph using causal interventions. Then, to fully capture the structural and feature information of the original graph, we perform enhanced condensation operations. Finally, through spectral-domain enhanced contrastive learning, we inject the causal-invariant features into the condensed graph, ensuring that the compressed graph retains the causal information of the original graph. Experimental results on five public datasets and our novel \textbf{FinReport} dataset demonstrate that TGCC achieves up to a 13.41\% improvement in cross-task and cross-domain complex scenarios compared to existing methods, and achieves state-of-the-art performance on 5 out of 6 datasets in the single dataset and task scenario.
\end{abstract}

\section{Introduction}
Graph Neural Networks (GNNs) \cite{wang2022heterogeneous,guo2023regraphgan,sun2024adaptive,lu2025progmlp} have garnered widespread attention for their exceptional ability to represent complex graph data and have been applied in many real-world scenarios, including social networks \cite{zhang2024information}, enterprise risk prediction \cite{zhao2022stock}, and recommender systems \cite{zhang2025unveiling}. It is widely believed that there exists a scaling law between dataset size and the capability of deep learning models \cite{kaplan2020scaling}. However, large-scale graph datasets also present significant challenges in terms of storage, processing, and computational resources. On one hand, specific applications such as neural architecture search \cite{zhang2024disentangled}, continual learning \cite{li2024matters}, and pre-training \cite{yu2025samgpt,liu2025graph} require repetitive training on these datasets, resulting in substantial computational costs. On the other hand, for users with limited computational resources, training models on large-scale graph datasets can be extremely time-consuming or even infeasible. Recently, several graph condensation (GC) methods \cite{sun2024gc} have been proposed to alleviate these issues.
 \begin{figure}[t]
     \centering
     \includegraphics[width=0.98\columnwidth]{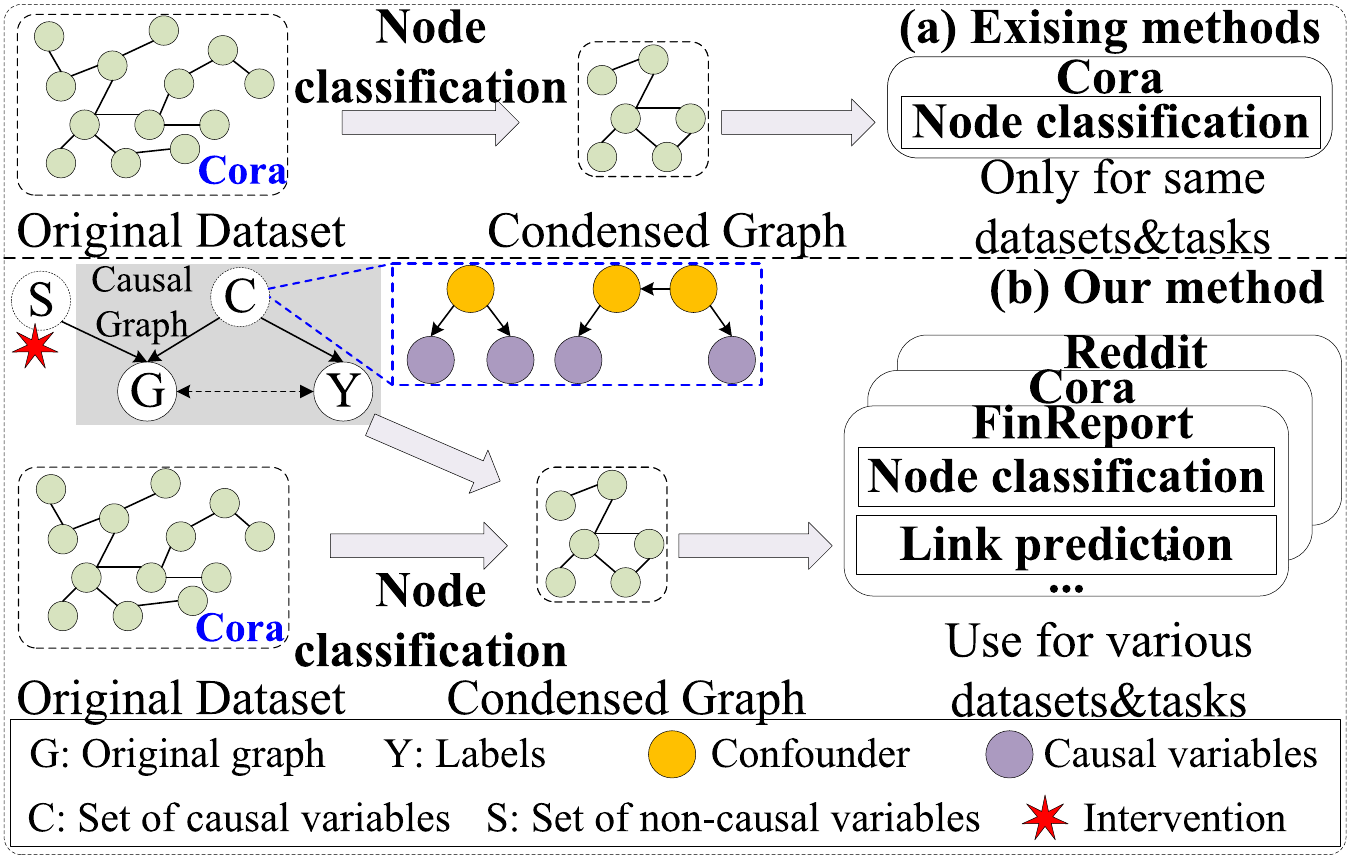}
     \caption{The pipeline of existing graph condensation methods and our TGCC method. The main differences between TGCC and existing methods lie in the extraction of causal-invariant features from the graph structure based on causal intervention to achieve transferable graph condensation.
}
     \label{fig:compare}
 \end{figure}

However, as shown in Fig. \ref{fig:compare}, most existing GC methods are designed based on statistical correlations to compress specific datasets, typically optimized for a single dataset and task. This limitation severely restricts the applicability of these methods in real-world scenarios, where users encounter new or custom data and tasks that differ significantly from the condensed graph datasets derived from publicly available datasets. 
% Such scenarios are much more common and challenging compared to applying methods on a few public graph datasets. 
Compared to applying GC methods on a few public graph datasets, such scenarios are much more common and challenging. While graph transfer learning \cite{zhuang2020comprehensive}, graph meta-learning \cite{fang2023community}, and graph foundation models \cite{liu2025graph} can partially address this issue, they rely on fixed model architectures or complex training strategies, and require substantial computational resources. Therefore, a natural question arises: \textit{Can the condensed graph dataset be used to train models that adapt to various tasks and different datasets?}

This paper focuses on the performance of models trained on condensed datasets when transferred to new datasets or tasks. Ideally, enhanced transferability would enable users to train models more efficiently, achieving better performance. However, current graph dataset condensation methods still face two major challenges.

Challenge 1: \textbf{Efficient and Fast Cross-task Adaptation}. Existing graph condensation algorithms typically condense, train, and test based on the same task, which prevents the trained models from generalizing effectively to other tasks. For example, using the Ogbn-arxiv dataset, we condense the dataset through node classification and then test the model’s performance on link prediction. The performance of existing methods is on average \textbf{3.2\%} lower than the ground truth (using only the simple 2-layer GCN), highlighting the difficulty of cross-task transferability.

Challenge 2: \textbf{Preservation of Causal Invariant Information}. Existing studies mainly perform condensation, training, and testing on the same dataset. The condensed graphs fail to capture the causal invariant information contained in the original graphs, leading to poor transferability of models trained on the condensed data. Using the Ogbn-arxiv as an example, we condense the dataset, then train a 2-layer GCN and generalize it to five different datasets by replacing the final layer with a new linear layer. However, the current methods still fail to outperform the simple GCN model, with an average performance loss of \textbf{9.8\%}.

To address the above challenges, we propose a causal-invariance-based and transferable graph dataset condensation framework, named \textbf{TGCC}. Unlike existing methods that typically focus on condensing a specific dataset for a single task, TGCC is designed to encourage the condensed graph to retain the most universal and information-rich patterns, thereby achieving better transferability across tasks and datasets. Specifically, to enhance transferability, TGCC first extracts causal-invariant features from the graph's structure. To better capture the structural and feature information of the original graph, TGCC performs contrastive condensation operations. Finally, through spectrum-level enhanced contrastive learning, the causal-invariant features are injected into the condensed graph, ensuring that the condensed graph retains the causal information of the original graph. Additionally, we construct a novel financial graph dataset, named \textbf{FinReport}, which captures the correspondence between corporate financial reports and analyst research reports, and release it as an open-source resource for the research community.

% We conduct empirical evaluations of TGCC on six real-world node-level graph datasets. The results show that in cross-dataset and cross-task scenarios, TGCC outperforms existing graph condensation methods by 2.5\% to 18.7\%. Even in traditional single-dataset and single-task settings, TGCC achieves state-of-the-art performance on all six datasets, demonstrating its versatility. 
In summary, the main contributions are stated as follows:

\noindent $\bullet$ We propose TGCC, a causal-invariant and transferable graph dataset condensation framework. To the best of our knowledge, TGCC is the first graph condensation method from a causal perspective that supports transferability.

\noindent $\bullet$ TGCC integrates spectral-domain intervention strategy and contrastive condensation strategy, leveraging contrastive learning to inject causal knowledge into the condensed graph, thereby enhancing its transferability across tasks and datasets.

\noindent $\bullet$ We construct a novel financial dataset, FinReport, capable of capturing the correspondence between corporate financial reports and company research reports, and we release it as an open-source resource for the research community.

\noindent $\bullet$ Extensive experiments on six real-world datasets demonstrate that TGCC achieves state-of-the-art performance in both single-task and cross-dataset/cross-task scenarios.
\section{Related Works}
\subsection{Graph Dataset Condensation}
The current GC methods are mainly divided into three categories: gradient matching, distribution matching, and trajectory matching. In recent years, several graph dataset condensation methods have been proposed to improve performance on single datasets and tasks \cite{xu2024survey,zhang2024navigating,gao2025graph,yu2025samgpt,xiao2025disentangled}. For instance, GCond \cite{jingraph} first introduced the gradient matching method for graph condensation. 
% GEOM \cite{zhang2024navigating} achieved lossless results (even outperforming training a GCN model solely on the original graph). 
ST-GCond \cite{yangst} offers a self-supervised and transferable graph dataset condensation method to address challenges in cross-task and cross-dataset scenarios. However, most existing methods are still designed for specific datasets and tasks. 
% \textcolor{red}{The only exception is ST-GCond}, 
The method most similar to ours, ST-GCond, which achieves transferability by pre-training multiple self-supervised model to extract fundamental information from the data. However, ST-GCond is costly and fails to extract causal-invariant features from graph data. Therefore, how to achieve transferable condensation remains an under-explored issue and requires further research.

 \subsection{Causal Inference in Graph Neural Networks}
Causality studies explore the relationships between variables \cite{pearl2016causal}, and have demonstrated many benefits in graph learning \cite{fan2022debiasing,dai2024comprehensive}. For example, 
% DIR-GNN \cite{wudiscovering} intervenes on the non-causal part to generate representations that uncover the reasoning process in the graph. 
% DisC \cite{fan2022debiasing} disentangles the graph into causal and bias subgraphs, alleviating biases in the dataset. 
RGCL \cite{li2022let} introduces an invariance perspective in self-supervised learning and proposes a method to preserve stable semantic information. GCIL \cite{mo2024graph} introduces spectral graph augmentation and proposes a new graph contrastive learning (GCL) method. UIL \cite{sui2025unified} provides a unified perspective on invariant graph learning, emphasizing both structural and semantic invariance principles to identify more robust stable features for graph classification. Unlike these studies, we explore a transferable graph condensation task from a causality perspective and propose a novel graph condensation framework based on causal theory.
 \section{The Proposed Model: TGCC}
 %  \begin{figure*}[htbp]
 %     \centering
 %     \includegraphics[width=0.90\linewidth]{fig/GC16.pdf}
 %     \caption{An illustrative diagram of the proposed TGCC framework.}
 %     \label{fig:framework}
 % \end{figure*}
 %  \subsection{Overall Framework}
  % In this paper, we aim to propose a transferable graph dataset condensation method to address cross-task and cross-dataset scenarios.
  \subsection{Overall Framework}
\begin{figure*}[t]
     \centering
     \includegraphics[width=0.88\linewidth]{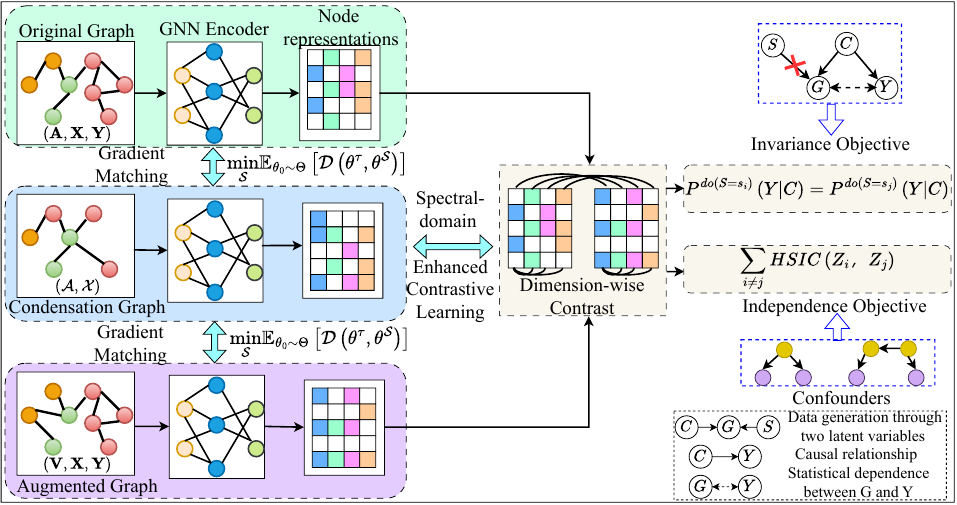}
     \caption{An illustrative diagram of the proposed TGCC framework.}
     \label{fig:framework}
 \end{figure*}

  Given a graph dataset 
$\mathcal{G}=\left ( X,A,Y\right )$, where 
$X\in \mathbb{R}^{N\times d}$
  represents the feature matrix, $A\in \mathbb{R}^{N\times N}$
  represents the adjacency matrix, $N$ is the number of nodes, and $d$ is the node feature dimension, our goal is to generate a smaller synthetic graph dataset $\mathcal{G}_{s}=\left ( X_{s},A_{s},Y_{s}\right )$, where the number of nodes $m\ll N$. We aim to train models on the synthetic graph $\mathcal{G}_{s}$ that achieve similar test performance to those trained on the original graph $\mathcal{G}$ for the same task, while preserving generalization ability when transferred to new datasets or tasks. 
  % Unlike existing condensation methods that are usually built upon data correlation \cite{yangst,gao2025rethinking} and are limited to specific tasks or datasets \cite{gao2025rethinking}, transferable graph dataset condensation seeks to broaden the applicability of the condensed graph.
  Therefore, we propose the TGCC framework as illustrated in Figure \ref{fig:framework}, which mainly consists of \textit{three modules}: Causal Invariant Feature Extraction, Graph Contrastive Condensation, and Spectral-domain Enhanced Contrastive Learning.
  \subsection{Causal Invariant Feature Extraction}
According to the study \cite{liu2022revisiting,mo2024graph}, perturbing the structure of the original graph alters the strength of frequency components in the graph spectrum, with the lowest-frequency information being approximately regarded as an invariant pattern across different views. Consistent with prior research \cite{mo2024graph}, we consider the low-frequency components in the graph as causal content, while the high-frequency components are treated as non-causal content. Therefore, we intervene on the non-causal variable $S$ in Figure 1 by disturbing the high-frequency information while keeping the low-frequency information unchanged.

Specifically, given the original adjacency matrix $A$, our goal is to obtain a new adjacency matrix $V$ via intervention, which constitutes the augmented graph ${\mathcal{G}}^{\prime}$, as follows:
  \begin{equation}
      V=A+\Delta_{A+}-\Delta_{A-}\ ,
  \end{equation}
  where: $\Delta_{A+}$
  denotes the edges to be added, $\Delta_{A-}$ denotes the edges to be deleted. We obtain $\Delta_{A+}$
  by maximizing the following objective function:
  \begin{equation}
  \begin{aligned}
      \mathcal{J}=&\overbrace{{\left \langle \Theta L, \Delta_{A+}\right\rangle}^{2}}^{\text{Matching} \;\text{term}}+
      \overbrace{\epsilon H\left ( \Delta_{A+}\right )}^{\text{Entropy}\; \text{regularization}}\\+& 
      \underbrace{\left \langle \Theta f, \Delta_{A+}\mathbf{1}_{n}-a\right\rangle+\left \langle \Theta l, {\Delta}^{T}_{A+}\mathbf{1}_{n}-b\right\rangle}_{\text{Lagrange} \;\text{constraint} \;\text{term}} \ ,
    \end{aligned}
\end{equation}
where $\Theta$ is a parameter updated during training, $L$ is the Laplacian matrix of the graph $\mathcal{G}$, $\epsilon$ is a weight parameter, $f$ and $l$ are Lagrange multipliers. $\left \langle P,Q\right \rangle=\textstyle\sum_{ij}{P}_{ij}{Q}_{ij}$. The vectors $a$ and $b$ are the node degree distributions. $H\left ( \cdotp \right )$ denotes the entropy regularization term, defined as:
$H\left ( P \right )=-\textstyle\sum_{i,j}P_{i,j}\left ( \mathrm{log}\left (P_{i,j} \right )-1\right )$. For the calculation of $\Delta_{A-}$, due to space limitations, please refer to \cite{liu2022revisiting}.

\subsubsection{Invariance Objective}Besides perturbing the non-causal factors $S$, it is also necessary to keep the causal factors $C$ unchanged. This process can be regarded as an intervention on $S$, enabling the model to learn invariant representations and thereby improve its generalization ability. Accordingly, the model should satisfy the following equation:
\begin{equation}\label{eq.inter}
    P^{do\left ( S=s_{i}\right )}\left ( Y\mid C\right )=P^{do\left ( S=s_{j}\right )}\left ( Y\mid C\right )\ ,
\end{equation}
where $do\left ( S=s\right )$ denotes the intervention on the non-causal factors $S$. In order to achieve the objective in Equation \ref{eq.inter}, based on causal inference theory \cite{pearl2009causality}, we can reformulate it as follows:
\begin{equation}
    CE\left ( C,S={s}_{i}\right )=CE\left ( C,S={s}_{j}\right )\ ,
\end{equation}
where $CE$ denotes the causal effect between variables. In other words, we need to capture the consistency between the node representations ${Z}^{A}$ and ${Z}^{V}$ 
% ${Z}^{A},{Z}^{V}\in \mathbb{R}^{N\times d}$ 
obtained by the encoder $f$ on $A$ and $V$, respectively. Following existing studies \cite{mo2024graph}, we assume that each dimension of the representation follows a Gaussian distribution, and we propose a dimension-level invariance objective, which aims to maintain consistency across views in each dimension. Specifically, by aligning the mean and standard deviation of each dimension of the representations. Formally, the learning objective is defined as:
\begin{equation}
\resizebox{0.86\linewidth}{!}{$
    \min_{g}{\displaystyle\sum_{i}{\left \| {Z}_{i}^{A}-{Z}_{i}^{V}\right \|}_{2}^{2}}, \:s.t.Std\left ( {Z}_{i}^{A}\right )=Std\left ( {Z}_{i}^{V}\right )=\lambda\ ,$}
\end{equation}
where ${Z}_{i}^{A}$ and ${Z}_{i}^{V}$ are the representations of the embedding matrices in the $i$-th dimension, and $Std$ is the standard deviation. The first term encourages the means of the embeddings to be aligned in the same dimension, while the second constraint pushes the standard deviations to approach the hyperparameter $\lambda$, thereby achieving dimension-level invariance.
 % \section{The Proposed Model: TGCC}
\subsubsection{Independence Objective}To consider a more common scenario where confounding variables exist in the causal graph, leading to spurious correlations between variables, we propose an independence objective to mitigate this challenge. This objective enforces mutual independence among different causal variables, thereby eliminating the correlations between them. In line with existing studies \cite{mo2024graph}, we adopt the Hilbert-Schmidt Independence Criterion (HSIC) to measure the independence between variables. When the value of HSIC is 0, it indicates that the two variables are independent. By minimizing the following objective function, we encourage the different dimensions in the representation matrix $Z$ to be mutually independent:
\begin{equation}
\resizebox{0.86\linewidth}{!}{$
    \displaystyle\sum_{i\ne j}HSIC\left ( {Z}_{i},{Z}_{j}\right )=\displaystyle\sum_{i\ne j}\frac{1}{\left ( {N-1}\right )^{2}}\mathrm{Tr}\left ( \mathbf{K}_{i}\mathbf{H}\mathbf{K}_{j}\mathbf{H}\right )\ ,$}
\end{equation}
where $\mathbf{H}$ is the centering matrix $\mathbf{I} - \frac{1}{N}\mathbf{1}\mathbf{1}^\top$, $\mathbf{I}$ is the identity matrix, ${Z}_{i}$ represents the $i$-th dimension of the embedding representation matrix, and $\mathbf{K}_{i}\in \mathbb{R}^{N \times N}$ and $\mathbf{K}_{j}\in \mathbb{R}^{N \times N}$ are the kernel matrices of $Z_i$ and $Z_j$, respectively.

% The kernel matrix $\mathbf{K}_i$ is used to compute the kernel function values between all pairs of samples for the corresponding variable. For example, $K^i_{a,b} = \kappa(Z_{i,a}, Z_{i,b})$, where $\kappa$ is the kernel function, $K^i_{a,b}$ represents the value at the $a$-th row and $b$-th column of $K_i$, and $Z_{i,a}$ denotes the value at the $a$-th row and $i$-th column in the representation matrix $Z$.

Please note that using complex kernel functions (e.g., Gaussian kernels) in HSIC to measure the independence between different dimensions can result in high spatial complexity, making it difficult to implement in scenarios with large sample sizes and high dimensions. Inspired by previous work \cite{mialon2022variance,mo2024graph}, we equivalently treat minimizing the HSIC between different dimensions as minimizing the sum of the off-diagonal elements of the covariance matrix. The proof for this conclusion is as follows:

Let the kernel function be defined as $k \left ( {Z}_{i,a},{Z}_{i,b}\right )=\psi\left ( {Z}_{i,a}\right )\psi\left ( {Z}_{i,b}\right )^{T}$, where $\psi$ is an elementwise mapping function. We denote the mapping of such projectors on the matrix 
$Z$ as $Q=\psi\left ( Z\right )=\left [\psi\left ( Z_{1}\right ),\dots ,\psi\left ( Z_{d}\right )  \right ]$. According to Lemma 1 in \citep{mialon2022variance}, we have:
\begin{equation}
\resizebox{0.90\linewidth}{!}{$
  \begin{aligned}
    HSIC\left ( {Z}_{i},{Z}_{j}\right )=&\frac{1}{{\left ( n-1\right )}^{2}}\mathrm{Tr}\left ( \psi\left ( {Z}_{i}\right )\psi\left ( {Z}_{i}\right )^{T}\mathbf{H}\psi\left ( {Z}_{j}\right )\psi\left ( {Z}_{j}\right )^{T}\mathbf{H}\right )\\=&\frac{1}{{\left ( n-1\right )}^{2}}{\left \| \psi\left ( {Z}_{i}\right )^{T}\mathbf{H}\psi\left ( {Z}_{j}\right )\right \|}_{F}^{2}={\left \| \mathrm{Cov}\left ( \psi\left ( {Z}_{i}\right ),\psi\left ( {Z}_{j}\right )\right )\right \|}_{F}^{2}\\=&{\left \| \mathrm{Cov}\left ( Q\right )_{\left ( i-1\right )L:iL,\left ( j-1\right )L:jL}\right \|}_{F}^{2}\ ,
    \end{aligned}$}
\end{equation}
where $\mathrm{Tr}$ denotes the trace of a matrix, and $\mathrm{Cov}$ represents the covariance between two variables. For complexity considerations and in line with existing studies \cite{mo2024graph}, we use a linear kernel, i.e., $g(X) = X$, in this case $Z = Q$, so we obtain:
\begin{equation}\label{eq.causal}
\resizebox{0.90\linewidth}{!}{$
    \textstyle\sum_{i\ne j}HSIC\left ( {Z}_{i},{Z}_{j}\right )=\textstyle\sum_{i\ne j}\mathrm{Cov}\left ( Q\right )_{i,j}^{2}=\textstyle\sum_{i\ne j}\mathrm{Cov}\left ( Z\right )_{i,j}^{2}\ .$}
\end{equation}
We convert the computation of HSIC values between different dimensions into the computation of covariance. By minimizing Eq. \ref{eq.causal}, the independence between different dimensions can be ensured.

In the this module, we further normalize the embedding matrix along each dimension, and denote the normalized node embeddings as $\bar{\mathbf{Z}}$. Note that since $\left \| \bar{\mathbf{Z}}\right \|^{2}=1$, ${\min}_{g}\textstyle\sum_{i}\left \| Z_{i}^{A}-Z_{i}^{V}\right \|^{2}_{2}$ can be equivalently replaced by ${\max}_{g}\textstyle\sum_{i}\bar{Z}_{i}^{A}\cdotp \bar{Z}_{i}^{V}$, where $\cdot$ denotes the inner product.
The $s_i$ represents the standard deviation of the $i$-th dimension before normalization. Minimizing $\sqrt{\left \| {s}_{i}^{A}-\lambda \right \|_{2}^{2}}$ encourages the standard deviation to approach the target value $\lambda$. The optimization objective is summarized as follows:
\begin{equation}
\resizebox{0.88\linewidth}{!}{$
\begin{aligned}\label{loss-causal}
   \mathcal{L}_{causal} = -&\alpha \sum_i \tilde{Z}_i^A \cdot \tilde{Z}_i^V 
+ \beta \sum_i \left\{ \sqrt{\|s_i^A - \lambda\|_2^2} + \sqrt{\|s_i^V - \lambda\|_2^2} \right\}\\ 
+&\gamma \sum_{i \ne j} \left (\text{Cov}(\tilde{Z}^A)^2_{i,j} + \text{Cov}(\tilde{Z}^V)^2_{i,j} \right )\ ,
\end{aligned}$}
\end{equation}
where $\alpha$, $\beta$, and $\gamma$ are hyperparameters. 
% And $\lambda$ is the desired standard deviation for each dimension.

\subsubsection{Theoretical analysis}We have the following theorem to depict the learning process of the TGCC.
\begin{theorem}\label{the1}
(Causal Invariance) Given adjacency matrix $A$ and the generated augmentation $V$ , the amplitudes of $i$-th frequency of $A$ and $V$ are ${\lambda }_{i}$ and ${\gamma }_{i}$, respectively. With the optimization of $\mathcal{L}_{causal}$, the following upper bound is established:
\begin{equation}
\resizebox{0.88\linewidth}{!}{$
\begin{aligned}
    \mathcal{L}_{causal}\leq \frac{1}{2}\displaystyle\sum_{i}{\theta }_{i}\left \| \lambda _{i}-\gamma  _{i}\right \|^{2}+\displaystyle\sum_{j}{\left [ {\left (\frac{1}{\sqrt{N}}\left \| {\lambda }_{j}^{A}\right \| -\lambda \right )}^{2}+{\left (\frac{1}{\sqrt{N}}\left \| {\gamma}_{j}^{V}\right \| -\lambda \right )}^{2}\right ]}\ ,
    \end{aligned}$}
\end{equation}
where ${\theta }_{i}$ is an adaptive weight of the $i$-th term.
\end{theorem}
Based on Theorem \ref{the1}, we theoretically prove that TGCC can capture the causal invariance information between $A$ and $V$. The proof is presented in \textbf{Appendix A}.

  \subsection{Graph Contrastive Condensation}
We take gradient matching as an example to obtain the condensed graph. It is worth noting that our method is compatible with any condensation approach. 
% Specifically, the objective of GC is defined as: $\min_{\mathcal{S}}{\mathcal{L}}^{\mathcal{T}}\left ( {\theta }^{\mathcal{S}}\right ) \;s.t.\;{\theta }^{\mathcal{S}}=\mathrm{arg}\min_{\theta }{{\mathcal{L}}^{\mathcal{S}}\left ( \theta \right )}$.
We formalize the condensation objective as:
\begin{equation}
 \min_{\mathcal{G}_{s}}\mathbb{E}_{\theta_{0}\backsim \Theta  }\left [ \mathcal{D}\left ( \theta^{\mathcal{G}},\theta^{\mathcal{G}_{s}}\right )\right ]\ ,   
\end{equation}
where $\theta_{0}$ is the initialization of both $\theta^{\mathcal{\mathcal{G}}}$
  and $\theta^{\mathcal{G}_{s}}$, $\mathcal{G}$ is the original graph, and $\Theta$ is a specific distribution used for relay model initialization. The expectation over $\theta_{0}$ aims to improve the robustness of the condensed data $\mathcal{G}_{s}$ against different parameter initializations \cite{lei2023comprehensive}. 
$\mathcal{D}\left ( \cdotp ,\cdotp \right )$ is a distance metric. The bi-level optimization objective is approximated by matching the model gradients at each training step. To fully capture both structural and feature information of the graph, we let the training trajectory on the condensed data mimic the training process on both the original and the augmented graphs. Therefore, the optimization objective for graph condensation, denoted as 
$\mathcal{L}_{\mathrm{cond}}$, is redefined as:
\begin{equation}
\resizebox{0.85\linewidth}{!}{$
  \begin{aligned}
    &\mathcal{L}_{\mathrm{cond}}={\mathbb{E}}_{{\theta }_{0}\backsim \Theta }\left [ \begin{aligned}&\displaystyle\sum_{t=1}^{T}\mathcal{D}\left ( \nabla_{\theta }{\mathcal{L}}^{\mathcal{G}}\left ( {\theta }_{t}\right ),\nabla_{\theta }{\mathcal{L}}^{\mathcal{G}_{s}}\left ( {\theta }_{t}\right )\right )\\&+\displaystyle\sum_{t=1}^{T} \mathcal{D}\left ( \nabla_{\theta }{\mathcal{L}}^{\mathcal{{G}}^{\prime}}\left ( {\theta }_{t}\right ),\nabla_{\theta }{\mathcal{L}}^{\mathcal{G}_{s}}\left ( {\theta }_{t}\right )\right )\end{aligned}\right ]\\ &\mathrm{s.t.} \;{\theta }_{t+1}=\mathrm{opt}\left ( {\mathcal{L}}^{\mathcal{G}_{s}}\left ( {\theta }_{t}\right )\right )\ ,
    \end{aligned}$}
\end{equation}
where $T$ is the number of training steps for the model, $\mathcal{G}^{\prime}=\left ( X,V,Y\right )$ is the augmented version of the original graph $\mathcal{G}$, $\mathrm{opt}\left ( \cdotp \right )$ is the model parameter optimizer and the parameter of relay model (e.g., GCN) is updated only on $\mathcal{G}_{s}$.
 \begin{figure}[t]
     \centering
     \includegraphics[width=0.82\columnwidth]{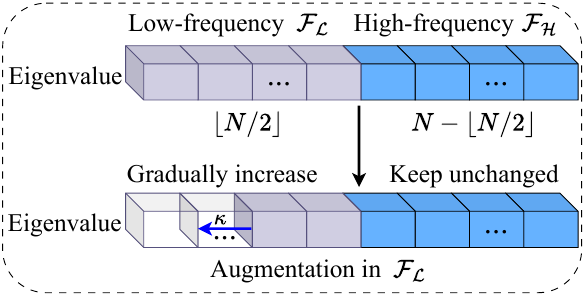}
     \caption{The generation of negative sample.}
     \label{fig:nega}
     \vspace{-0.8em}
 \end{figure}
  \subsection{Spectral-domain Enhanced Contrastive Learning}
To inject causally invariant information from the original graph structure into the condensed graph and train a model with stronger generalization ability, we adopt a spectral-domain enhanced contrastive learning strategy. Previous studies \cite{liu2022revisiting} theoretically proves that GCL can learn invariant information between two augmented graphs, and this invariance is concentrated in the low-frequency components. Additionally, our exploratory experiments also confirm the same conclusion (see \textbf{Appendix F}). Therefore, we retain the high-frequency information and perturb the low-frequency components to construct negative samples. The specific construction process is illustrated in Figure \ref{fig:nega} and the corresponding expression is as follows:
\begin{equation}\label{eq:des}
\resizebox{0.98\linewidth}{!}{$
  \begin{aligned}
    \hat{\mathcal{L}}=&{\lambda }_{\left ( 1-\kappa\right ) * N/2}\mathbf{u}_{\left ( 1-\kappa\right ) * N/2}\mathbf{u}^{\top }_{\left ( 1-\kappa\right )* N/2}+\cdots +{\lambda }_{ \left \lfloor N/2 \right \rfloor-1 }\mathbf{u}_{\left \lfloor N/2 \right \rfloor-1}\mathbf{u}^{\top }_{\left \lfloor N/2 \right \rfloor-1}\\+&{\lambda }_{\left \lfloor N/2 \right \rfloor}\mathbf{u}_{\left \lfloor N/2 \right \rfloor}\mathbf{u}^{\top}_{\left \lfloor N/2 \right \rfloor}+\cdots +{\lambda }_{N}\mathbf{u}_{N}\mathbf{u}^{\top }_{N}\ ,
    \end{aligned}$}
\end{equation}
where $\mathbf{u}_{i}$ is the eigenvector corresponding to eigenvalue ${\lambda }_{i}$,  and $\kappa$ is the proportion of eigenvalues added in descending order. $\hat{\mathcal{L}}$ is the symmetric normalized graph Laplacian, from which the adjacency matrix of negative samples can be further obtained based on the degree matrix.

The core idea of GCL is to train the embedding of a target graph in one augmented view to be as close as possible to the embedding of its positive sample in another augmented view, while being far from those of its negative samples. Models constructed in this way can effectively distinguish between similar and dissimilar graphs. In this work, we adopt the InfoNCE loss as the optimization objective:
\begin{equation}
\resizebox{0.86\linewidth}{!}{$
    \mathcal{L}_{\mathrm{InfoNCE}}=\log \frac{\exp\left ( \varphi \left ( {h}^{{V}_{i}},{h}^{{V}_{j}}\right )/t\right )}{\exp\left (\varphi \left ( {h}^{{V}_{i}},{h}^{{V}_{j}}\right )/t\right )+\displaystyle\sum_{m\ne i}\exp\left ( \varphi \left ( {h}^{{V}_{i}},{h}^{{V}_{m}}\right )/t\right )}\ ,$}
\end{equation}
where ${h}^{{V}_{i}}$ and ${h}^{{V}_{j}}$ represent the graph embeddings obtained by applying a readout function to the node feature matrices of the condensed graph and the causally invariant features, respectively. $\varphi$ is the similarity measure function, such as cosine similarity, and $t$ is the temperature parameter. ${h}^{{V}_{m}}$ denotes the graph embedding corresponding to the constructed negative sample.
\subsection{Optimization Objective}
% In the Causal Invariant Feature Extraction module, we further normalize the embedding matrix along each dimension, and denote the normalized node embeddings as $\bar{\mathbf{Z}}$. Note that since $\left \| \bar{\mathbf{Z}}\right \|^{2}=1$, ${\min}_{g}\textstyle\sum_{i}\left \| Z_{i}^{A}-Z_{i}^{B}\right \|^{2}_{2}$ can be equivalently replaced by ${\max}_{g}\textstyle\sum_{i}\bar{Z}_{i}^{A}\cdotp \bar{Z}_{i}^{B}$, where $\cdot$ denotes the inner product.
% The $s_i$ represents the standard deviation of the $i$-th dimension before normalization. Minimizing $\sqrt{\left \| {s}_{i}^{A}-\lambda \right \|_{2}^{2}}$ encourages the standard deviation to approach the target value $\lambda$.

The overall optimization objective of our proposed method TGCC is summarized as follows:
\begin{equation}
\resizebox{0.7\linewidth}{!}{$
\label{all.loss}
    \mathcal{L} =\mathcal{L}_{causal}+\delta\mathcal{L}_{\mathrm{InfoNCE}}+\eta\mathcal{L}_{\mathrm{cond}}\ ,$}
\end{equation}
where $\delta$ and $\eta$ are hyperparameters that control the importance of each term in the loss function. The complexity analysis of the algorithm can be found in \textbf{Appendix D}.

%   \subsection{Complexity Analysis}
% The overall pipeline of TGCC consists of three modules: Causal Invariant Feature Extraction, Graph Contrastive Compression, and Spectral-domain Enhanced Contrastive Learning. 

% We analyze the time complexity of each module separately as follows: (1) Causal Invariant Feature Extraction. The complexity of the first term of the loss corresponding to this module is $\mathcal{O}(Nd)$, the second term is $\mathcal{O}(Nd)$, and the third term is $\mathcal{O}(d^2)$, resulting in a total complexity of $\mathcal{O}(Nd^2)$. (2) Graph Contrastive Compression. In the comparison compression module, the time complexity of node feature propagation is $\mathcal{O}\left ( Nd
% +E\right )$, and the time complexity of gradient matching is $\mathcal{O}\left ( T\left ( Nd
% +E\right )\right )$. Therefore, the overall time complexity of the module is: $\mathcal{O}\left ( T\left ( Nd
% +E\right )\right )$, where $T$ is the number of gradient matching iterations, $E$ is the number of edges. (3) Spectral-domain Enhanced Contrastive Learning. The time complexity of constructing negative samples is $\mathcal{O}(d^{2})$. 

\begin{table}[t]
  \centering
  % \vspace{-0.6em}
   % \vspace{-0.6em}
  \resizebox{0.44\textwidth}{!}{
  % \scriptsize
    \begin{tabular}{l|c|c|c|c}
    \toprule
    Datasets & \#Classes&\#Nodes&\#Edges&\#Feature\\
    \midrule
   \textbf{Cora} & 7& 2,708&5,429&1,433  \\
   \textbf{Citeseer} &6&3,327&4,732&3,703 \\
   \textbf{Flickr} &7&89,250&899,756&500  \\
   \textbf{Ogbn-Arxiv} &40&169,343&1,166,243&128\\
   \textbf{Reddit} & 210&232,965&57,307,946&602\\
   \midrule
    \textbf{FinReport (our)} & 7&4,992&3,568& 384\\
    \bottomrule
    \end{tabular}
    }
% \vspace{-0.1em}
%   \caption{Inference time per instance (s).}
  \caption{Statistics of datasets.}
  \label{fig:dataset}%
  \vspace{-0.8em}
\end{table}
 \begin{table*}[htbp]
\centering
  % \scriptsize
\resizebox{0.92\textwidth}{!}{
\begin{tabular}{l|c|ccc|ccccc|c|c}
 \toprule
\multicolumn{1}{l|}{\textbf{Datasets}} &
\multicolumn{1}{c|}{\textbf{Ratio ($r$)
}} &
\multicolumn{1}{c}{\textbf{Random}} & \multicolumn{1}{c}{\textbf{Herding
}} & \multicolumn{1}{c|}{\textbf{K-Center}}& \multicolumn{1}{c}
{\textbf{DosCond}} & \multicolumn{1}{c}
{\textbf{GCond}} & \multicolumn{1}{c}
{\textbf{SGDD}}&
\multicolumn{1}{c}{\textbf{GDEM}}&
\multicolumn{1}{c|}{\textbf{CGC}}& \multicolumn{1}{c|}
% {\textbf{DosCond}}& \multicolumn{1}{c|}
% {\textbf{KiDD}}& \multicolumn{1}{c}
{\textbf{TGCC}}&\multicolumn{1}{c}{\textbf{Whole
Dataset}} \\ 
 \midrule
\multicolumn{1}{c|}{\multirow{3}[2]{*}{Cora}}     &1.3\%  & 58.0$\pm$\textit{\scriptsize 1.4}&54.0$\pm$\textit{\scriptsize 0.7}&58.0$\pm$\textit{\scriptsize 1.1}&66.0$\pm$\textit{\scriptsize 2.4}&63.6$\pm$\textit{\scriptsize 1.0}& 63.7$\pm$\textit{\scriptsize 1.1} &\underline{67.0}$\pm$\textit{\scriptsize 1.2}&64.4$\pm$\textit{\scriptsize 1.1}&\textbf{67.6}$\pm$\textit{\scriptsize 1.8}&\multicolumn{1}{c}{\multirow{3}[2]{*}{78.5$\pm$\textit{\scriptsize 1.1}}}                \\
&2.6\%&55.0$\pm$\textit{\scriptsize 1.4}&55.0$\pm$\textit{\scriptsize 1.8}&56.0$\pm$\textit{\scriptsize 1.3}&64.3$\pm$\textit{\scriptsize 1.3}&66.4$\pm$\textit{\scriptsize 0.9}&59.7$\pm$\textit{\scriptsize 1.3}&\underline{67.8}$\pm$\textit{\scriptsize 4.5}&60.1$\pm$\textit{\scriptsize 0.1}&\textbf{69.8}$\pm$\textit{\scriptsize 0.4}&\\
&5.2\%&57.0$\pm$\textit{\scriptsize 1.8}&56.0$\pm$\textit{\scriptsize 1.9}&58.0$\pm$\textit{\scriptsize 1.3}&61.5$\pm$\textit{\scriptsize 0.5}&\underline{61.7}$\pm$\textit{\scriptsize 0.7}&55.0$\pm$\textit{\scriptsize 0.6}&54.4$\pm$\textit{\scriptsize 1.0}&60.1$\pm$\textit{\scriptsize 0.1}&\textbf{65.7}$\pm$\textit{\scriptsize 2.1}&\\
\midrule
\multicolumn{1}{c|}{\multirow{3}[2]{*}{Citeseer}} &0.9\% &52.0$\pm$\textit{\scriptsize 1.3}&52.0$\pm$\textit{\scriptsize 1.6}&55.0$\pm$\textit{\scriptsize 0.6}&60.0$\pm$\textit{\scriptsize 5.3}& 60.3$\pm$\textit{\scriptsize 1.5}&67.1$\pm$\textit{\scriptsize 3.4} &\underline{72.3}$\pm$\textit{\scriptsize 1.8} &61.4$\pm$\textit{\scriptsize 2.5}&\textbf{74.1}$\pm$\textit{\scriptsize 0.2}&\multicolumn{1}{c}{\multirow{3}[2]{*}{81.2$\pm$\textit{\scriptsize 1.2}}}   \\
&1.8\%&52.0$\pm$\textit{\scriptsize 0.5}&52.0$\pm$\textit{\scriptsize 1.7}&54.0$\pm$\textit{\scriptsize 0.9}&50.8$\pm$\textit{\scriptsize 1.5}&64.0$\pm$\textit{\scriptsize 2.3}&55.1$\pm$\textit{\scriptsize 2.1}&\underline{72.1}$\pm$\textit{\scriptsize 1.0}&64.8$\pm$\textit{\scriptsize 0.9}&\textbf{74.6}$\pm$\textit{\scriptsize 0.9}&\\
&3.6\%&54.0$\pm$\textit{\scriptsize 0.2}&53.0$\pm$\textit{\scriptsize 0.8}&53.0$\pm$\textit{\scriptsize 1.0}&57.5$\pm$\textit{\scriptsize 3.2}&\underline{68.7}$\pm$\textit{\scriptsize 2.6}&53.6$\pm$\textit{\scriptsize 0.8}&53.5$\pm$\textit{\scriptsize 1.0} &59.0$\pm$\textit{\scriptsize 1.2}&\textbf{69.4}$\pm$\textit{\scriptsize 1.2}&\\
\midrule
\multicolumn{1}{c|}{\multirow{3}[2]{*}{Ogbn-arxiv}}&0.05\% &69.7$\pm$\textit{\scriptsize 1.0}                      &70.3$\pm$\textit{\scriptsize 1.1}&\underline{70.4}$\pm$\textit{\scriptsize 0.7}&68.8$\pm$\textit{\scriptsize 0.6}& 69.2$\pm$\textit{\scriptsize 1.1}&68.2$\pm$\textit{\scriptsize 1.2}&70.2$\pm$\textit{\scriptsize 0.4}&70.2$\pm$\textit{\scriptsize 0.6}&\textbf{71.0}$\pm$\textit{\scriptsize 0.3}&\multicolumn{1}{c}{\multirow{3}[2]{*}{74.1$\pm$\textit{\scriptsize 0.2}}}  \\
&0.25\%&70.5$\pm$\textit{\scriptsize 0.3}&70.6$\pm$\textit{\scriptsize 0.5}&70.5$\pm$\textit{\scriptsize 0.3}&70.1$\pm$\textit{\scriptsize 0.8}&70.7$\pm$\textit{\scriptsize 0.4}&68.8$\pm$\textit{\scriptsize 0.8}&\textbf{72.5}$\pm$\textit{\scriptsize 0.3}&70.7$\pm$\textit{\scriptsize 0.2}&\underline{70.9}$\pm$\textit{\scriptsize 0.7}&\\
&0.5\%&70.8$\pm$\textit{\scriptsize 0.4}&70.9$\pm$\textit{\scriptsize 0.5}&70.1$\pm$\textit{\scriptsize 0.5}&71.5$\pm$\textit{\scriptsize 0.6}&71.5$\pm$\textit{\scriptsize 0.3}&69.9$\pm$\textit{\scriptsize 0.4}&71.4$\pm$\textit{\scriptsize 0.2}&\underline{71.6}$\pm$\textit{\scriptsize 0.0}&\textbf{72.8}$\pm$\textit{\scriptsize 0.5}&\\
\midrule
\multicolumn{1}{c|}{\multirow{3}[2]{*}{flickr}} &0.1\%&57.2$\pm$\textit{\scriptsize 2.7}&54.7$\pm$\textit{\scriptsize 2.9}&57.4$\pm$\textit{\scriptsize 1.3}&55.3$\pm$\textit{\scriptsize 1.6}&\underline{70.1}$\pm$\textit{\scriptsize 0.4} &55.4$\pm$\textit{\scriptsize 1.5}&53.3$\pm$\textit{\scriptsize 0.8}&54.8$\pm$\textit{\scriptsize 2.6}&\textbf{70.6}$\pm$\textit{\scriptsize 0.5}&\multicolumn{1}{c}{\multirow{3}[2]{*}{74.9$\pm$\textit{\scriptsize 0.1}}}\\
&0.5\%&69.6$\pm$\textit{\scriptsize 1.0}&68.7$\pm$\textit{\scriptsize 1.5}&67.2$\pm$\textit{\scriptsize 0.7}&61.6$\pm$\textit{\scriptsize 2.2}&\underline{70.0}$\pm$\textit{\scriptsize 0.4}&62.7$\pm$\textit{\scriptsize 1.3}&53.3$\pm$\textit{\scriptsize 0.7}&52.6$\pm$\textit{\scriptsize 2.1}&\textbf{71.0}$\pm$\textit{\scriptsize 0.6}&\\
&1\%&70.1$\pm$\textit{\scriptsize 0.4}&\underline{70.3}$\pm$\textit{\scriptsize 0.7}&\textbf{70.9}$\pm$\textit{\scriptsize 0.6}&64.9$\pm$\textit{\scriptsize 1.2}&69.7$\pm$\textit{\scriptsize 0.5}&66.5$\pm$\textit{\scriptsize 1.1}&54.0$\pm$\textit{\scriptsize 1.2}&53.4$\pm$\textit{\scriptsize 1.8}&68.5$\pm$\textit{\scriptsize 0.7}&\\
   \midrule
\multicolumn{1}{c|}{\multirow{3}[2]{*}{reddit}}&0.05\%&59.0$\pm$\textit{\scriptsize 1.7}&60.1$\pm$\textit{\scriptsize 1.3}&59.6$\pm$\textit{\scriptsize 2.0}&55.1$\pm$\textit{\scriptsize 1.3}&\underline{64.9}$\pm$\textit{\scriptsize 2.4}&56.1$\pm$\textit{\scriptsize 0.8}&61.2$\pm$\textit{\scriptsize 2.8}&64.6$\pm$\textit{\scriptsize 1.1}&\textbf{73.6}$\pm$\textit{\scriptsize 1.1}&\multicolumn{1}{c}{\multirow{3}[2]{*}{81.6$\pm$\textit{\scriptsize 0.1}}}\\
&0.1\%&61.2$\pm$\textit{\scriptsize 1.3}&61.0$\pm$\textit{\scriptsize 2.5}&59.6$\pm$\textit{\scriptsize 2.0}&57.8$\pm$\textit{\scriptsize 1.6}&62.1$\pm$\textit{\scriptsize 1.4}&58.4$\pm$\textit{\scriptsize 1.1}&62.0$\pm$\textit{\scriptsize 2.6}&\underline{63.9}$\pm$\textit{\scriptsize 2.1}&\textbf{69.9}$\pm$\textit{\scriptsize 1.0}&\\
&0.2\%&63.4$\pm$\textit{\scriptsize 1.1}&\underline{65.8}$\pm$\textit{\scriptsize 2.1}&60.6$\pm$\textit{\scriptsize 1.9}&63.5$\pm$\textit{\scriptsize 1.5}&63.7$\pm$\textit{\scriptsize 4.5}&63.5$\pm$\textit{\scriptsize 1.9}&63.9$\pm$\textit{\scriptsize 2.2}&62.3$\pm$\textit{\scriptsize 1.7}&\textbf{72.4}$\pm$\textit{\scriptsize 0.9}&\\
\midrule
\multicolumn{1}{c|}{\multirow{3}[2]{*}{FinReport}} &2.02\%&70.6$\pm$\textit{\scriptsize 1.1}&70.7$\pm$\textit{\scriptsize 0.9}&71.1$\pm$\textit{\scriptsize 0.8}&72.2$\pm$\textit{\scriptsize 1.5}&\underline{73.8}$\pm$\textit{\scriptsize 1.1}&70.9$\pm$\textit{\scriptsize 1.6}&72.6$\pm$\textit{\scriptsize 1.1}&73.1$\pm$\textit{\scriptsize 0.1}&\textbf{74.6}$\pm$\textit{\scriptsize 0.8} &\multicolumn{1}{c}{\multirow{3}[2]{*}{77.0$\pm$\textit{\scriptsize 0.3}}}\\
&4.05\%&72.3$\pm$\textit{\scriptsize 0.5}&71.9$\pm$\textit{\scriptsize 0.9}&71.5$\pm$\textit{\scriptsize 0.3}&\underline{74.7}$\pm$\textit{\scriptsize 0.6}&73.8$\pm$\textit{\scriptsize 0.9}&72.4$\pm$\textit{\scriptsize 1.7}&67.4$\pm$\textit{\scriptsize 3.8}&73.2$\pm$\textit{\scriptsize 0.2}&\textbf{75.4}$\pm$\textit{\scriptsize 0.6}&\\
&6.07\%&72.4$\pm$\textit{\scriptsize 0.4}&71.2$\pm$\textit{\scriptsize 0.6}&71.5$\pm$\textit{\scriptsize 1.1}&\underline{74.6}$\pm$\textit{\scriptsize 0.7}&72.8$\pm$\textit{\scriptsize 0.8}&74.5$\pm$\textit{\scriptsize 1.7}&72.8$\pm$\textit{\scriptsize 1.9}&71.9$\pm$\textit{\scriptsize 0.0}&\textbf{74.9}$\pm$\textit{\scriptsize 1.1}&\\
\bottomrule
\end{tabular}}
\caption{
Link prediction performance (Accuracy\% ± std) comparison under the cross-task and single-dataset setting. The best results are highlighted in \textbf{bold}, and the runner-up results are \textbf{underlined}. $r = m/N$. \textbf{Code}: \url{https://github.com/HYJ9999/TGCC.}}
\label{fig-single-link}
\vspace{-1em}
\end{table*}
 \section{Experiments}
 \subsection{Experimental Setup}
 \subsubsection{Datasets}
 We evaluate our method on five public datasets (Cora \cite{kipf2017semi},
Citeseer \cite{kipf2017semi}, Ogbn-arxiv \cite{hu2020open}, Reddit \cite{hamilton2017inductive}, Flickr \cite{zenggraphsaint}), and our constructed dataset: \textbf{FinReport}. For the node classification task, we follow the settings from GCond \cite{jingraph}. For the link prediction task, we follow the public split of the dataset. The statistics of the datasets are provided in Table \ref{fig:dataset}. We report the details of the datasets in \textbf{Appendix B}.
  \subsubsection{Baselines}
To evaluate the performance of TGCC, we conduct experiments using four types of baselines, including: (1) traditional core-set methods including Random, Herding \cite{welling2009herding}, K-Center \cite{farahani2009facility}, (2) gradient matching methods including DosCond \cite{jin2022automated}, GCond \cite{jingraph} and SGDD \cite{yang2023does}, (3) trajectory matching methods including SFGC \cite{zheng2023structure} and GEOM \cite{zhang2024navigating} , and (4) distribution matching methods including GDEM \cite{liu2024graph} and CGC \cite{gao2025rethinking}. The details of evaluated methods are in \textbf{Appendix C}.
\subsubsection{Implementation Details} 
% \textit{In the condensing stage}, for each dataset, the compression ratio $r$ is defined as the number of nodes in the compressed graph, $r = \frac{N}{m}$, where $0 < r < 1$. For each dataset, we compress 5 graphs using different random seeds and report the average performance. \textit{In the testing stage}, we choose the appropriate testing paradigm based on the specific scenario:
For each dataset, we perform condensation on 5 graphs using different random seeds and report the average performance.
\textit{Single-dataset and single-task scenario}: We train the model using the condensed graph and evaluate it on the original graph. \textit{Cross-dataset and cross-task scenario}: We train the model using the condensed graph, and then train a linear classifier using the downstream data. The parameter settings are provided in \textbf{Appendix E}.

\subsection{Experiments Results}
To thoroughly evaluate the performance of TGCC, we conduct experiments in \underline{four scenarios}: (1) single-task and single-dataset (see \textbf{Appendix G1}); (2) cross-dataset; (3) cross-task; and (4) cross-dataset and cross-task.

 \textbf{Performance Comparison on Cross-task Scenario.}
  Following existing studies \cite{yangst}, we primarily focus on using node classification tasks for condensation and evaluate link prediction performance on the test set. As shown in Table \ref{fig-single-link}, TGCC achieves either the best or second-best performance across all datasets compared to the baselines. Notably, on the Reddit, our method outperforms the second-best model, GCond, by 13.41\%, indicating that the compressed graph effectively preserves causal knowledge beneficial for cross-task scenarios. However, on the Flickr ($r$= 1\%), TGCC performs relatively poorly. A potential reason is the presence of complex confounding relationships and latent variables in the data, which may hinder the accurate extraction of causally invariant features.
\begin{table*}[t]
\centering
\resizebox{0.92\textwidth}{!}{
\begin{tabular}{l|ccc|ccc|ccc|ccc|ccc}
\toprule
\multirow{4}{*}{Methods} & \multicolumn{15}{c}{Ogbn-arxiv $\rightarrow$ Target datasets} \\
\cmidrule(lr){2-16}
 & \multicolumn{3}{c|}{Cora} & \multicolumn{3}{c|}{Citeseer} & \multicolumn{3}{c|}{Flickr} & \multicolumn{3}{c|}{Reddit}& \multicolumn{3}{c}{FinReport} \\
\cmidrule(lr){2-4} \cmidrule(lr){5-7} \cmidrule(lr){8-10} \cmidrule(lr){11-13}\cmidrule(lr){14-16}
 & 0.05\% & 0.25\% & 0.5\% & 0.05\% & 0.25\% & 0.5\% & 0.05\% & 0.25\% & 0.5\% & 0.05\% & 0.25\% & 0.5\% & 0.05\% & 0.25\% & 0.5\% \\
\midrule
Random & 51.5&58.5& 58.0&44.0& 56.7& 56.3& 44.6&\underline{44.9}&\underline{44.9}&76.1& 84.6&85.5&65.1&\underline{66.1}&65.9\\
Herding & 48.0& 56.6& 57.8& 48.6& 54.6& 56.1& 44.4& 44.7& \textbf{45.0}& 73.6& 83.3& 85.2&62.1&66.0&64.1\\
GCond &57.0& 56.7& 54.3&55.7&52.1&50.4&44.6&44.6&44.5& 85.7& 84.5& 85.0&66.6&65.9&\underline{66.4}\\
SGDD & 53.6& \underline{58.6}& \underline{59.0}& \underline{55.8}& \underline{57.0}& \underline{57.5}&44.6&44.6&44.6& \underline{85.8}& \underline{85.0}& \underline{85.9}&\underline{67.0}&65.6&66.3\\
SFGC & 55.3& 57.5& \underline{59.0}& 50.0& 49.9& 55.4& \underline{44.9}& \underline{44.9}& 44.8& 81.4& 84.5& \textbf{86.2}&65.0&65.7&65.5\\
GDEM & 56.7& 58.0& 42.9& 54.2& 46.6& 40.4& \underline{44.9}& 44.2& 43.1& 85.2& 82.1&67.3&62.7&62.8&60.9\\
GEOM & \underline{57.2}& 55.6& 58.2& 44.4& 54.1& 56.7& 44.6& 44.8& \underline{44.9}& 81.5& 84.5&\underline{85.9}&65.8&65.7&66.3\\
CGC & 53.9& 58.3& 56.7& 51.8& 53.6& 52.0& 44.8& 44.8& 44.5& 84.9& 84.2&84.8&65.2&66.0&66.3\\
\midrule
TGCC & \textbf{59.9}& \textbf{61.1}& \textbf{60.3}& \textbf{56.5}& \textbf{57.4}& \textbf{57.9}&\textbf{45.3}& \textbf{45.1}& 44.8&\textbf{86.5}& \textbf{85.7}& 85.1&\textbf{67.9}&\textbf{66.3}&\textbf{67.2}\\
\bottomrule
\end{tabular}}
\caption{Node classification performance comparison (Accuracy\%) under the cross-dataset scenario.}
\label{tab:cross-node-ogbn}
% \vspace{-0.5em}
\end{table*}

\begin{table*}[t]
\centering
\resizebox{0.93\textwidth}{!}{
\begin{tabular}{l|cccccc|cccccc|cccccc}
\toprule
\multirow{5}{*}{Methods} & \multicolumn{18}{c}{Flickr $\rightarrow$ Target datasets} \\
\cmidrule(lr){2-19}
 & \multicolumn{6}{c|}{Cora} & \multicolumn{6}{c|}{Citeseer}& \multicolumn{6}{c}{Reddit}\\
\cmidrule(lr){2-7} \cmidrule(lr){8-13} \cmidrule(lr){14-19} 
 & \multicolumn{2}{c}{0.1\%} & \multicolumn{2}{c}{0.5\%} & \multicolumn{2}{c|}{1.0\%} & \multicolumn{2}{c}{0.1\%} & \multicolumn{2}{c}{0.5\%} & \multicolumn{2}{c|}{1.0\%}& \multicolumn{2}{c}{0.1\%} & \multicolumn{2}{c}{0.5\%} & \multicolumn{2}{c}{1.0\%}\\
 \cmidrule(lr){2-7} \cmidrule(lr){8-13} \cmidrule(lr){14-19} 
    & AUC& AP& AUC& AP & AUC& AP& AUC& AP& AUC& AP& AUC& AP& AUC& AP& AUC& AP& AUC& AP\\
\midrule
Random &51.2&50.6& 54.2&52.2& 54.7&52.5&54.5&52.4&62.2&56.9&62.8&57.4& 55.3&52.9& \underline{63.3}&\underline{57.7}& 61.3&55.8\\
GCond &\underline{55.6}&\underline{53.1}& \underline{57.3}&\underline{53.1}& \underline{57.2}&53.7&\underline{61.4}&\underline{55.7}&\underline{63.4}&\underline{57.8}&\underline{63.0}&\underline{57.5}& \underline{58.4}&\underline{53.5}& 62.5&57.2& \underline{61.8}&\underline{56.4}\\
SGDD & 51.7&50.9& 54.4&52.3& 57.0&\underline{53.8}& 53.1&51.6& 57.4&54.0& 61.6&56.5&50.9&50.4&55.3&52.8&59.1&55.5\\
DosCond & 54.3&52.3& 53.3&51.7& 55.1&52.7& 55.9&53.2& 56.0&53.2& 61.7&56.7& 52.8&51.4& 55.7&53.0&60.8&56.1\\
GDEM & 52.7&51.4& 51.3&50.7& 53.0&51.5& 53.8&52.0& 52.2&51.1& 53.3&51.7& 52.4&51.2& 50.7&50.3&51.6&50.8\\
CGC & 51.2&50.6& 50.6&50.3& 50.5&50.3& 50.7&50.4& 51.2&50.6& 51.6&50.8& 50.6&50.3& 50.0&50.0&50.1&50.1\\
\midrule
TGCC & \textbf{60.5}&\textbf{55.9}&\textbf{60.4}&\textbf{55.8}& \textbf{59.7}&\textbf{55.4}& \textbf{65.5}&\textbf{59.2}& \textbf{67.3}&\textbf{60.4}& \textbf{65.5}&\textbf{59.2}&\textbf{62.6}&\textbf{57.3}& \textbf{66.9}&\textbf{60.2}& \textbf{63.8}&\textbf{58.0}\\
\bottomrule
\end{tabular}}
\caption{Link prediction performance comparison under the cross-task and cross-dataset. AP stands for Average Precision score.
}
\label{tab:cross-link-flickr}
\vspace{-0.8em}
\end{table*}

 \begin{table}[t]
  \centering
  % \vspace{-0.6em}
   % \vspace{-0.6em}
  \resizebox{0.88\columnwidth}{!}{
    \begin{tabular}{l|ccccc|c}
    \toprule
    Methods & GCN&SAGE&SGC&APPNP&Cheby&AVG\\
    \midrule
   \textbf{GCond} &87.4& 86.2&87.0&84.6&\underline{68.3}&82.7 \\
   \textbf{SGDD} &86.0&83.1&87.5&62.2&59.3&75.6\\
   \textbf{DosCond}&49.5&50.4&53.1&37.6&33.0&44.7\\
   \textbf{SFGC} &78.3&81.1&89.2&\textbf{88.0}&58.4&79.0\\
    \textbf{CGC} &\underline{88.0}&\underline{86.9}&\textbf{90.5}&\underline{87.9}&61.0&\underline{82.9}\\
   \textbf{TGCC} &\textbf{88.5}&\textbf{88.3}&\underline{89.5}&87.5&\textbf{69.4}&\textbf{84.6}\\
    \bottomrule
    \end{tabular}%
    }
% \vspace{-0.1em}
%   \caption{Inference time per instance (s).}
  \caption{The generalizability of GC methods on Reddit ($r$ = 0.05\%). AVG indicates the average value.}
  \label{fig:general}%
  % \vspace{-0.5em}
\end{table}%
  \textbf{Performance Comparison on Cross-dataset Scenario.} We compare TGCC with baseline methods under the cross-dataset transfer learning setting. We use Ogbn-arxiv as the source dataset and test the condensed graph on five other target datasets. The results are shown in Table \ref{tab:cross-node-ogbn}. We observe that TGCC achieves the best performance in most cases. These improvements demonstrate the effectiveness of incorporating causal invariant information from the graph structure into universal knowledge extraction, allowing the condensed graph to benefit various downstream datasets. Furthermore, TGCC achieves better results on target datasets, such as the FinReport dataset, compared to using the GCN model alone. Therefore, downstream users can achieve similar test performance to expensive GCN models with significantly lower computational costs by training models on the condensed graph and using a simple linear classifier. Additionally, we use FinReport as the source dataset and test the condensed graph on four other target datasets, with results presented in Table 10 in \textbf{Appendix G2}, further demonstrating the effectiveness of our method.
  
    \textbf{Performance Comparison on Cross-task and Cross-dataset Scenario.}
 We compare TGCC with major baseline methods under the cross-dataset and cross-task setting. We use Flickr as the source dataset and evaluate the link prediction performance of the condensed graph on three other target datasets. The results are shown in Table \ref{tab:cross-link-flickr}. It can be observed that TGCC achieves the best performance. Notably, on the Reddit, our model achieves an AUC improvement of 7.2\% and an AP improvement of 7.1\%. These improvements further demonstrate the effectiveness of causally invariant information in graph structures, enabling the condensed graph to better support various downstream tasks across different datasets, and also offering a new perspective for the development and training of graph foundation models.
 % The complete experimental results are provided in Table \ref{tab:cross-link-ogbn-1} in the \textbf{Appendix C}.
 More results and analysis can be found in \textbf{Appendix G3}.
 
  \textbf{Generalizability Comparison.}
  To compare the generalizability of different GNN architectures, we evaluated the node classification performance of GC methods under different GNN models, including GCN, SGC, SAGE \cite{hamilton2017inductive}, APPNP \cite{gasteigerpredict}, and Cheby \cite{defferrard2016convolutional}. We take Reddit as an example to evaluate the performance of different GNN architectures. Table \ref{fig:general} presents the detailed accuracy results. We observe that GCN and SGC outperform other GNNs, as they adopt the same propagation mechanism used in the feature propagation process during condensation. In addition, TGCC achieves a significant improvement over other baseline methods, indicating that our proposed method can effectively capture causal information in graph data and contribute to better performance on downstream tasks.

\begin{figure}[t]
     \centering
     \includegraphics[width=0.865\columnwidth]{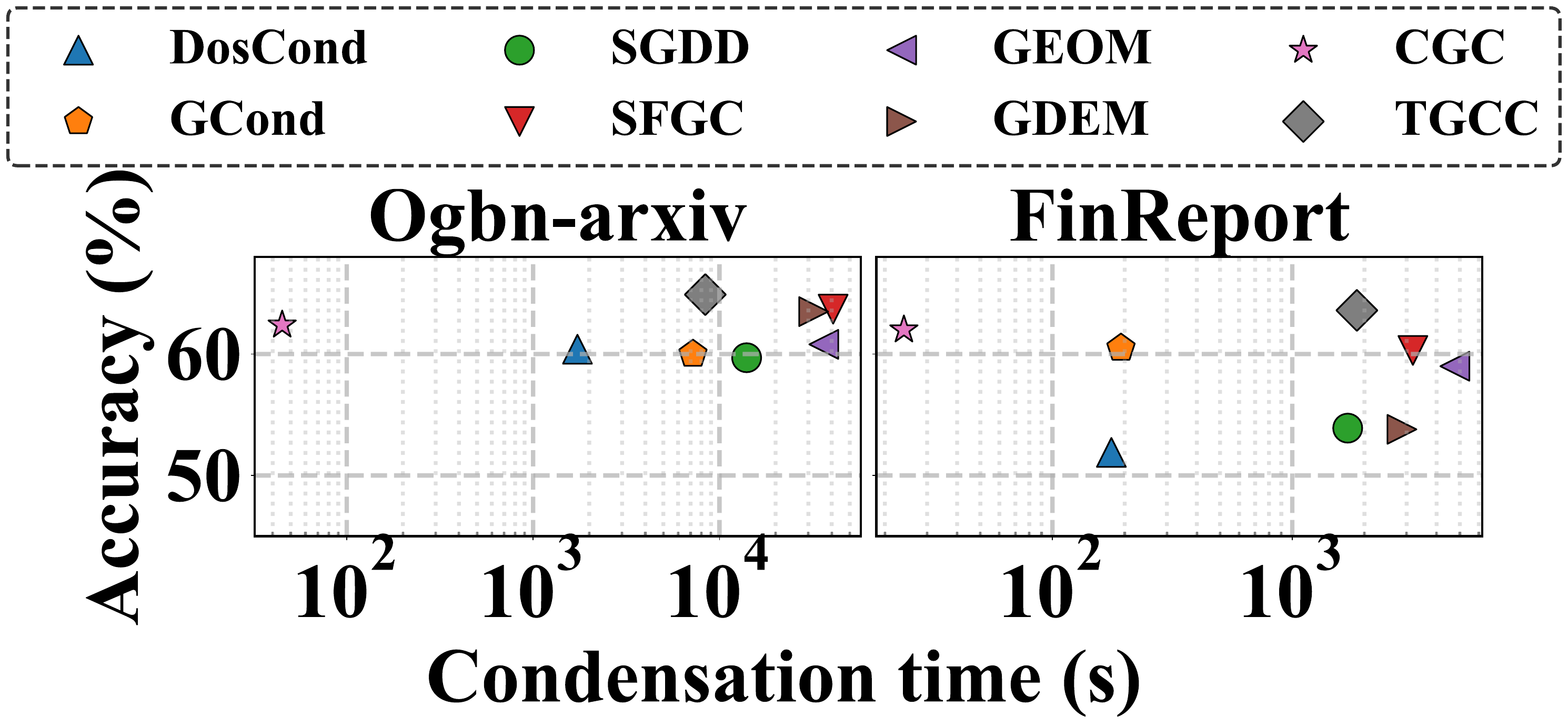}
	\caption{The accuracy and condensation time of GC methods on Ogbn-Arxiv ($r$ = 0.5\%) and FinReport ($r$ = 2.02\%).}
     \label{fig:time}
     \vspace{-1.2em}
 \end{figure}

\begin{table}[t]
\centering
\resizebox{0.99\columnwidth}{!}{
\begin{tabular}{l|ccc|ccc}
\toprule
\multirow{2}{*}{Methods} & \multicolumn{3}{c|}{\makecell{Flickr}} & \multicolumn{3}{c}{\makecell{Citeseer}} \\
\cmidrule(lr){2-4} \cmidrule(lr){5-7}
& $r$=0.1\% & $r$=0.5\% & $r$=1.0\% & $r$=0.9\% & $r$=1.8\%& $r$=3.6\%\\
\midrule
% Random & 59.1$_{\pm1.7}$ & 58.1$_{\pm1.9}$ & 68.3$_{\pm1.4}$ & 69.0$_{\pm1.0}$ \\
% Herding & 47.3$_{\pm2.5}$ & 50.2$_{\pm3.0}$ & 74.7$_{\pm2.4}$ & 75.3$_{\pm0.8}$ \\
% SGDD & 63.6$_{\pm2.0}$ & 67.0$_{\pm2.2}$ & 85.2$_{\pm1.6}$ & 84.1$_{\pm1.8}$ \\
% GEOM & 66.0$_{\pm1.1}$ & 65.9$_{\pm2.1}$ & 87.1$_{\pm1.0}$ & 87.5$_{\pm0.8}$ \\
% \midrule
w/o CIFE & 46.8& 46.8& 46.9& 70.1&69.6&69.5\\
w/o GCC & 45.7& 46.8& 47.2& 70.6&69.3&70.2\\
w/o ECL & 43.5& 46.8&47.2&70.3&70.5&68.1\\
\midrule
\textbf{TGCC}& \textbf{50.2}& \textbf{50.2}& \textbf{50.3}& \textbf{73.0}&\textbf{73.3}&\textbf{72.8}\\
\bottomrule
\end{tabular}}
\caption{Node classification performance in ablation study.}
\label{tab:ablation}
\vspace{-1.2em}
\end{table}
\textbf{Efficiency Comparison.}
To clearly compare the efficiency of these methods, Figure \ref{fig:time} shows the accuracy of GC methods versus their condensation time on the Ogbn-Arxiv and FinReport. It can be observed
that our TGCC achieves the highest test accuracy, and is 3 times and 2 times faster than the SOTA baselines SFGC and GEOM, respectively.

  \textbf{Ablation Study and Sensitivity Analysis.}
  To investigate the impact of different modules in our method, we conduct an ablation study on node classification tasks using Flickr and Citeseer as examples.
  % , as shown in Table \ref{tab:ablation}.
 The variants include: \textbf{w/o CIFE} (excluding the causal invariant feature extraction module), \textbf{w/o GCC} (excluding the graph condensation module), and \textbf{w/o ECL} (excluding the spectral-domain enhanced contrastive Learning). From Table \ref{tab:ablation}, we observe that TGCC achieves the best performance when all components are included. Removing any single module leads to a performance drop, which validates the effectiveness of jointly considering all three components. Due to space limitations, additional ablation studies and detailed parameter sensitivity analyses can be found in \textbf{Appendix G4} and \textbf{G5}, respectively.
% \begin{figure}[t]
% 	\centering
%     % 删除子标题, 只保留图片
%     \includegraphics[width=0.234\textwidth]{fig/ogbn_time_0.5_14.pdf}
%     \hfill
%     \includegraphics[width=0.234\textwidth]{fig/finance_time_0.5_28.pdf}
% 	\caption{The accuracy and condensation time of GC methods on Ogbn-Arxiv ($r$ = 0.5\%) and FinReport ($r$ = 2.02\%).}
% 	\label{fig:time}
%     % \vspace{-1.4em}
% \end{figure}
% \subsection{Efficiency Comparison}
% To facilitate a clearer comparison of the efficiency of these methods, Figure \ref{fig:time} shows the accuracy of GC methods versus their condensation time on the Ogbn-Arxiv and FinReport dataset. From the figure, it can be seen that our proposed TGCC achieves the highest test accuracy, and is 3 times and 2 times faster than the state-of-the-art baseline methods SFGC and GEOM, respectively.
 \section{Conclusion}
We propose TGCC, a novel causal-invariance-based and transferable graph dataset condensation framework. 
% Unlike existing methods that focus on condensation for a single task and dataset, our approach creates condensed datasets with better transferability, enhancing the downstream model's ability when applied to various new datasets and tasks. 
We first extract causal-invariant knowledge from the graph structure based on causal interventions. Then, through contrastive condensation operations, we extract structural and feature information from the original graph. Finally, we guide the condensation process by using spectral domain enhancement contrastive loss, injecting the causal information. Extensive experiments demonstrate the effectiveness of TGCC.

\section{Acknowledgements}
This research is partially supported by the CAAI-Ant Research Fund (CAAI-MYJJ 2024-03), Xiangjiang Laboratory (25XJ02002), as well as funding from the National Natural Science Foundation of China (62376228, 62376227, 62472240, 62394322, U22B2031, 72401060, 72442025), the Science and Technology Innovation Program of Hunan Province (2024RC4008), the China Postdoctoral Science Foundation (2025M770766), Sichuan Provincial Postdoctoral Research Project Special Funding (TB2025043), Liaoning Provincial Natural Science Foundation Doctoral Research Start-up Project (2025-BS-0833), the Taishan Scholar Foundation of Shandong Province (tstp20250724), the Beijing National Research Center for Information Science and Technology (BNR2025RC01010), and Sichuan Science and Technology Program (2023NSFSC0032).  Carl
Yang is not supported by any funds from China.

In addition, we would like to express our sincere gratitude to the Ant Group team for their outstanding collaboration and unwavering support. From data collection, cleaning, and structuring to more complex tasks such as manual/automated collaborative labeling, system development, and model algorithm design, the contributions of the Ant team have been indispensable at every stage. Their professionalism in project management, resource coordination, and technical problem-solving has played a pivotal role in ensuring the smooth progress of the project. It is through such highly efficient collaboration and deep technical expertise that the project has achieved significant progress. We look forward to continuing our fruitful partnership with Ant Group in the future and exploring new opportunities for cutting-edge research and technological innovation together.
\bibliography{aaai2026}

@inproceedings{kipf2017semi,
  title={Semi-Supervised Classification with Graph Convolutional Networks},
  author={Kipf, Thomas N and Welling, Max},
  booktitle={International Conference on Learning Representations},
  year={2017}
}

@inproceedings{hu2020open,
  title={Open graph benchmark: Datasets for machine learning on graphs},
  author={Hu, Weihua and Fey, Matthias and Zitnik, Marinka and Dong, Yuxiao and Ren, Hongyu and Liu, Bowen and Catasta, Michele and Leskovec, Jure},
  booktitle={Advances in neural information processing systems},
  volume={33},
  pages={22118--22133},
  year={2020}
}

@inproceedings{hamilton2017inductive,
  title={Inductive representation learning on large graphs},
  author={Hamilton, Will and Ying, Zhitao and Leskovec, Jure},
  booktitle={Advances in neural information processing systems},
  volume={30},
  year={2017}
}

@inproceedings{zenggraphsaint,
  title={GraphSAINT: Graph Sampling Based Inductive Learning Method},
  author={Zeng, Hanqing and Zhou, Hongkuan and Srivastava, Ajitesh and Kannan, Rajgopal and Prasanna, Viktor},
  booktitle={International Conference on Learning Representations},
  year={2020}
}

@inproceedings{jingraph,
  title={Graph Condensation for Graph Neural Networks},
  author={Jin, Wei and Zhao, Lingxiao and Zhang, Shichang and Liu, Yozen and Tang, Jiliang and Shah, Neil},
  booktitle={International Conference on Learning Representations},
  year={2022}
}

@inproceedings{welling2009herding,
  title={Herding dynamical weights to learn},
  author={Welling, Max},
  booktitle={Proceedings of the 26th annual international conference on machine learning},
  pages={1121--1128},
  year={2009}
}

@inproceedings{jin2022automated,
  title={AUTOMATED SELF-SUPERVISED LEARNING FOR GRAPHS},
  author={Jin, Wei and Liu, Xiaorui and Zhao, Xiaoyu and Ma, Yao and Shah, Neil and Tang, Jiliang},
  booktitle={10th International Conference on Learning Representations, ICLR 2022},
  year={2022}
}

@article{yang2023does,
  title={Does graph distillation see like vision dataset counterpart?},
  author={Yang, Beining and Wang, Kai and Sun, Qingyun and Ji, Cheng and Fu, Xingcheng and Tang, Hao and You, Yang and Li, Jianxin},
  journal={Advances in Neural Information Processing Systems},
  volume={36},
  pages={53201--53226},
  year={2023}
}

@article{zheng2023structure,
  title={Structure-free graph condensation: From large-scale graphs to condensed graph-free data},
  author={Zheng, Xin and Zhang, Miao and Chen, Chunyang and Nguyen, Quoc Viet Hung and Zhu, Xingquan and Pan, Shirui},
  journal={Advances in Neural Information Processing Systems},
  volume={36},
  pages={6026--6047},
  year={2023}
}

@inproceedings{zhang2024navigating,
  title={Navigating Complexity: Toward Lossless Graph Condensation via Expanding Window Matching},
  author={Zhang, Yuchen and Zhang, Tianle and Wang, Kai and Guo, Ziyao and Liang, Yuxuan and Bresson, Xavier and Jin, Wei and You, Yang},
  booktitle={International Conference on Machine Learning},
  pages={60379--60395},
  year={2024},
  organization={PMLR}
}

@article{guo2023regraphgan,
  title={RegraphGAN: A graph generative adversarial network model for dynamic network anomaly detection},
  author={Guo, Dezhi and Liu, Zhaowei and Li, Ranran},
  journal={Neural Networks},
  volume={166},
  pages={273--285},
  year={2023},
  publisher={Elsevier}
}

@article{wang2022heterogeneous,
  title={Heterogeneous network representation learning approach for ethereum identity identification},
  author={Wang, Yixian and Liu, Zhaowei and Xu, Jindong and Yan, Weiqing},
  journal={IEEE Transactions on Computational Social Systems},
  volume={10},
  number={3},
  pages={890--899},
  year={2022},
  publisher={IEEE}
}

@article{sun2024adaptive,
  title={Adaptive attention-based graph representation learning to detect phishing accounts on the ethereum blockchain},
  author={Sun, Haojie and Liu, Zhaowei and Wang, Shenqiang and Wang, Haiyang},
  journal={IEEE Transactions on Network Science and Engineering},
  volume={11},
  number={3},
  pages={2963--2975},
  year={2024},
  publisher={IEEE}
}

@article{lu2025progmlp,
  title={ProGMLP: A Progressive Framework for GNN-to-MLP Knowledge Distillation with Efficient Trade-offs},
  author={Lu, Weigang and Guan, Ziyu and Zhao, Wei and Yang, Yaming and Sun, Yujie and Liang, Zheng and Zhan, Yibing and Tao, Dapeng},
  journal={arXiv preprint arXiv:2507.19031},
  year={2025}
}

@book{farahani2009facility,
  title={Facility location: concepts, models, algorithms and case studies},
  author={Farahani, Reza Zanjirani and Hekmatfar, Masoud},
  year={2009},
  publisher={Springer Science \& Business Media}
}

@book{pearl2016causal,
  title={Causal inference in statistics: A primer},
  author={Pearl, Judea and Glymour, Madelyn and Jewell, Nicholas P},
  year={2016},
  publisher={John Wiley \& Sons}
}

@book{pearl2009causality,
  title={Causality},
  author={Pearl, Judea},
  year={2009},
  publisher={Cambridge university press}
}

@article{fan2022debiasing,
  title={Debiasing graph neural networks via learning disentangled causal substructure},
  author={Fan, Shaohua and Wang, Xiao and Mo, Yanhu and Shi, Chuan and Tang, Jian},
  journal={Advances in Neural Information Processing Systems},
  volume={35},
  pages={24934--24946},
  year={2022}
}

@inproceedings{li2022let,
  title={Let invariant rationale discovery inspire graph contrastive learning},
  author={Li, Sihang and Wang, Xiang and Zhang, An and Wu, Yingxin and He, Xiangnan and Chua, Tat-Seng},
  booktitle={International conference on machine learning},
  pages={13052--13065},
  year={2022},
  organization={PMLR}
}

@article{xu2024survey,
  title={A survey on graph condensation},
  author={Xu, Hongjia and Zhang, Liangliang and Ma, Yao and Zhou, Sheng and Zheng, Zhuonan and Jiajun, Bu},
  journal={arXiv preprint arXiv:2402.02000},
  year={2024}
}

@article{gao2025graph,
  title={Graph condensation: A survey},
  author={Gao, Xinyi and Yu, Junliang and Chen, Tong and Ye, Guanhua and Zhang, Wentao and Yin, Hongzhi},
  journal={IEEE Transactions on Knowledge and Data Engineering},
  year={2025},
  publisher={IEEE}
}

@inproceedings{mo2024graph,
  title={Graph contrastive invariant learning from the causal perspective},
  author={Mo, Yanhu and Wang, Xiao and Fan, Shaohua and Shi, Chuan},
  booktitle={Proceedings of the AAAI conference on artificial intelligence},
  volume={38},
  number={8},
  pages={8904--8912},
  year={2024}
}

@inproceedings{gasteigerpredict,
  title={Predict then Propagate: Graph Neural Networks meet Personalized PageRank},
  author={Gasteiger, Johannes and Bojchevski, Aleksandar and G{\"u}nnemann, Stephan},
  booktitle={International Conference on Learning Representations},
year={2019}
}

@inproceedings{gao2025rethinking,
  title={Rethinking and accelerating graph condensation: A training-free approach with class partition},
  author={Gao, Xinyi and Ye, Guanhua and Chen, Tong and Zhang, Wentao and Yu, Junliang and Yin, Hongzhi},
  booktitle={Proceedings of the ACM on Web Conference 2025},
  pages={4359--4373},
  year={2025}
}

@inproceedings{zhang2024disentangled,
  title={Disentangled continual graph neural architecture search with invariant modular supernet},
  author={Zhang, Zeyang and Wang, Xin and Qin, Yijian and Chen, Hong and Zhang, Ziwei and Chu, Xu and Zhu, Wenwu},
  booktitle={Forty-first International Conference on Machine Learning},
  year={2024}
}

@inproceedings{yu2025samgpt,
  title={Samgpt: Text-free graph foundation model for multi-domain pre-training and cross-domain adaptation},
  author={Yu, Xingtong and Gong, Zechuan and Zhou, Chang and Fang, Yuan and Zhang, Hui},
  booktitle={Proceedings of the ACM on Web Conference 2025},
  pages={1142--1153},
  year={2025}
}

@inproceedings{xiao2025disentangled,
  title={Disentangled condensation for large-scale graphs},
  author={Xiao, Zhenbang and Wang, Yu and Liu, Shunyu and Hu, Bingde and Wang, Huiqiong and Song, Mingli and Zheng, Tongya},
  booktitle={Proceedings of the ACM on Web Conference 2025},
  pages={4494--4506},
  year={2025}
}

@inproceedings{fang2023community,
  title={Community search: a meta-learning approach},
  author={Fang, Shuheng and Zhao, Kangfei and Li, Guanghua and Yu, Jeffrey Xu},
  booktitle={2023 IEEE 39th International Conference on Data Engineering (ICDE)},
  pages={2358--2371},
  year={2023},
  organization={IEEE}
}

@article{zhuang2020comprehensive,
  title={A comprehensive survey on transfer learning},
  author={Zhuang, Fuzhen and Qi, Zhiyuan and Duan, Keyu and Xi, Dongbo and Zhu, Yongchun and Zhu, Hengshu and Xiong, Hui and He, Qing},
  journal={Proceedings of the IEEE},
  volume={109},
  number={1},
  pages={43--76},
  year={2020},
  publisher={Ieee}
}

@article{sun2024gc,
  title={Gc-bench: An open and unified benchmark for graph condensation},
  author={Sun, Qingyun and Chen, Ziying and Yang, Beining and Ji, Cheng and Fu, Xingcheng and Zhou, Sheng and Peng, Hao and Li, Jianxin and Yu, Philip S},
  journal={Advances in Neural Information Processing Systems},
  volume={37},
  pages={37900--37927},
  year={2024}
}

@article{liu2025graph,
  title={Graph foundation models: Concepts, opportunities and challenges},
  author={Liu, Jiawei and Yang, Cheng and Lu, Zhiyuan and Chen, Junze and Li, Yibo and Zhang, Mengmei and Bai, Ting and Fang, Yuan and Sun, Lichao and Yu, Philip S and others},
  journal={IEEE Transactions on Pattern Analysis and Machine Intelligence},
  year={2025},
  publisher={IEEE}
}

@article{li2024matters,
  title={What matters in graph class incremental learning? An information preservation perspective},
  author={Li, Jialu and Wang, Yu and Zhu, Pengfei and Lin, Wanyu and Hu, Qinghua},
  journal={Advances in Neural Information Processing Systems},
  volume={37},
  pages={26195--26223},
  year={2024}
}

@article{kaplan2020scaling,
  title={Scaling laws for neural language models},
  author={Kaplan, Jared and McCandlish, Sam and Henighan, Tom and Brown, Tom B and Chess, Benjamin and Child, Rewon and Gray, Scott and Radford, Alec and Wu, Jeffrey and Amodei, Dario},
  journal={arXiv preprint arXiv:2001.08361},
  year={2020}
}

@inproceedings{zhang2024information,
  title={Information diffusion meets invitation mechanism},
  author={Zhang, Shiqi and Sun, Jiachen and Lin, Wenqing and Xiao, Xiaokui and Huang, Yiqian and Tang, Bo},
  booktitle={Companion Proceedings of the ACM Web Conference 2024},
  pages={383--392},
  year={2024}
}

@article{zhao2022stock,
  title={Stock movement prediction based on bi-typed hybrid-relational market knowledge graph via dual attention networks},
  author={Zhao, Yu and Du, Huaming and Liu, Ying and Wei, Shaopeng and Chen, Xingyan and Zhuang, Fuzhen and Li, Qing and Kou, Gang},
  journal={IEEE transactions on knowledge and data engineering},
  volume={35},
  number={8},
  pages={8559--8571},
  year={2022},
  publisher={IEEE}
}

@inproceedings{zhang2025unveiling,
  title={Unveiling Contrastive Learning's Capability of Neighborhood Aggregation for Collaborative Filtering},
  author={Zhang, Yu and Zhang, Yiwen and Zhang, Yi and Sang, Lei and Yang, Yun},
  booktitle={Proceedings of the 48th International ACM SIGIR Conference on Research and Development in Information Retrieval},
  pages={1985--1994},
  year={2025}
}

@inproceedings{defferrard2016convolutional,
  title={Convolutional neural networks on graphs with fast localized spectral filtering},
  author={Defferrard, Micha{\"e}l and Bresson, Xavier and Vandergheynst, Pierre},
  booktitle={Proceedings of the 30th International Conference on Neural Information Processing Systems},
  pages={3844--3852},
  year={2016}
}

@article{mialon2022variance,
  title={Variance covariance regularization enforces pairwise independence in self-supervised representations},
  author={Mialon, Gr{\'e}goire and Balestriero, Randall and LeCun, Yann},
  journal={arXiv preprint arXiv:2209.14905},
  year={2022}
}

@article{lei2023comprehensive,
  title={A comprehensive survey of dataset distillation},
  author={Lei, Shiye and Tao, Dacheng},
  journal={IEEE Transactions on Pattern Analysis and Machine Intelligence},
  volume={46},
  number={1},
  pages={17--32},
  year={2023},
  publisher={IEEE}
}

@article{liu2022revisiting,
  title={Revisiting graph contrastive learning from the perspective of graph spectrum},
  author={Liu, Nian and Wang, Xiao and Bo, Deyu and Shi, Chuan and Pei, Jian},
  journal={Advances in Neural Information Processing Systems},
  volume={35},
  pages={2972--2983},
  year={2022}
}

@inproceedings{yangst,
  title={ST-GCond: Self-supervised and Transferable Graph Dataset Condensation},
  author={Yang, Beining and Sun, Qingyun and Ji, Cheng and Fu, Xingcheng and Li, Jianxin},
  booktitle={The Thirteenth International Conference on Learning Representations},
  year={2025}
}

@inproceedings{liu2024graph,
  title={Graph data condensation via self-expressive graph structure reconstruction},
  author={Liu, Zhanyu and Zeng, Chaolv and Zheng, Guanjie},
  booktitle={Proceedings of the 30th ACM SIGKDD Conference on Knowledge Discovery and Data Mining},
  pages={1992--2002},
  year={2024}
}

@article{dai2024comprehensive,
  title={A comprehensive survey on trustworthy graph neural networks: Privacy, robustness, fairness, and explainability},
  author={Dai, Enyan and Zhao, Tianxiang and Zhu, Huaisheng and Xu, Junjie and Guo, Zhimeng and Liu, Hui and Tang, Jiliang and Wang, Suhang},
  journal={Machine Intelligence Research},
  volume={21},
  number={6},
  pages={1011--1061},
  year={2024},
  publisher={Springer}
}

@inproceedings{sui2025unified,
  title={A Unified Invariant Learning Framework for Graph Classification},
  author={Sui, Yongduo and Sun, Jie and Wang, Shuyao and Liu, Zemin and Cui, Qing and Li, Longfei and Wang, Xiang},
  booktitle={KDD},
  year={2025}
}
\end{document}